\documentclass{article}

\usepackage{arxiv}

\usepackage[utf8]{inputenc} 
\usepackage[T1]{fontenc}    
\usepackage{hyperref}       
\usepackage{url}            
\usepackage{booktabs}       
\usepackage{amsfonts}       
\usepackage{nicefrac}       
\usepackage{microtype}      
\usepackage{lipsum}		
\usepackage{graphicx}
\usepackage{natbib}
\usepackage{doi}
\usepackage{amsmath,subcaption, textcomp, colortbl, enumitem}
\usepackage[linesnumbered,ruled,vlined]{algorithm2e}
\definecolor{lightgray}{gray}{0.95}

\title{A Circular Construction Product Ontology for End-of-Life Decision-Making}

\author{
\begin{tabular}{ccc}  
\href{https://orcid.org/0000-0003-2418-9314}
{\includegraphics[scale=0.06]{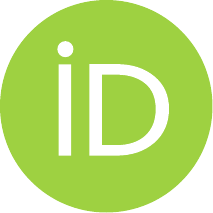}\hspace{1mm}\textbf{Kwabena Adu-Duodu}} &
\href{https://orcid.org/0000-0002-1647-2251}
{\includegraphics[scale=0.06]{orcid.pdf}\hspace{1mm}\textbf{Stanly Wilson}} &
\href{https://orcid.org/0000-0002-1647-2251}
{\includegraphics[scale=0.06]{orcid.pdf}\hspace{1mm}\textbf{Yinhao Li}} \\
\textnormal{Department of Computer Science} & \textnormal{Department of Computer Science} & \textnormal{Department of Computer Science} \\
\textnormal{Newcastle University, UK} & \textnormal{Newcastle University, UK} & \textnormal{Newcastle University, UK} \\
\texttt{K.Adu-Duodu2@newcastle.ac.uk} & \texttt{S.W.Palathingal2@newcastle.ac.uk} & \texttt{Yinhao.Li@newcastle.ac.uk}
\end{tabular} \\
\\[0.001cm]
\begin{tabular}{ccc}  
\href{https://orcid.org/0009-0007-1971-3704}
{\includegraphics[scale=0.06]{orcid.pdf}\hspace{1mm}\textbf{Aanuoluwapo Oladimeji}} &
\href{https://orcid.org/0009-0008-4797-5856}
{\includegraphics[scale=0.06]{orcid.pdf}\hspace{1mm}\textbf{Talea Huraysi}} &
\href{https://orcid.org/0000-0001-7829-2240}
{\includegraphics[scale=0.06]{orcid.pdf}\hspace{1mm}\textbf{Masoud Barati}} \\
\textnormal{Computer and Information Sciences} & \textnormal{Department of Computer Science} & \textnormal{School of Information Technology} \\
\textnormal{Northumbria University, UK} & \textnormal{Newcastle University, UK} & \textnormal{Carleton University, Canada} \\
\texttt{Oladimeji.aanu@gmail.com} & \texttt{t.h.a.huraysi2@newcastle.ac.uk} & \texttt{Masoud.Barati@carleton.ca}
\end{tabular}
\\
\\[0.001cm]
\begin{tabular}{ccc}  
\href{https://orcid.org/0000-0002-0190-3346}
{\includegraphics[scale=0.06]{orcid.pdf}\hspace{1mm}\textbf{Charith Perera}} &
\href{https://orcid.org/0000-0002-8346-7962}
{\includegraphics[scale=0.06]{orcid.pdf}\hspace{1mm}\textbf{Ellis Solaiman}} &
\href{https://orcid.org/0000-0003-3597-2646}
{\includegraphics[scale=0.06]{orcid.pdf}\hspace{1mm}\textbf{Omer Rana}} \\
\textnormal{Computer Science and Informatics} & \textnormal{Department of Computer Science} & \textnormal{Computer Science and Informatics} \\
\textnormal{Cardiff University, UK} & \textnormal{Newcastle University, UK} & \textnormal{Cardiff University, UK} \\
\texttt{PereraC@cardiff.ac.uk} & \texttt{Ellis.Solaiman@newcastle.ac.uk} & \texttt{RanaOF@cardiff.ac.uk}
\end{tabular} 
\\
\\[0.01cm] 
\begin{tabular}{cc}  
\href{https://orcid.org/0000-0002-6610-1328}
{\includegraphics[scale=0.06]{orcid.pdf}\hspace{1mm}\textbf{Rajiv Ranjan}} &
\href{https://orcid.org/0000-0001-7060-4211}
{\includegraphics[scale=0.06]{orcid.pdf}\hspace{1mm}\textbf{Tejal Shah}} \\
\textnormal{Department of Computer Science} & \textnormal{Department of Computer Science} \\
\textnormal{Newcastle University, UK} & \textnormal{Newcastle University, UK} \\
\texttt{Raj.Ranjan@newcastle.ac.uk} & \texttt{Tejal.Shah@newcastle.ac.uk}
\end{tabular}
}

\renewcommand{\shorttitle}{\textit{arXiv} Template}

\hypersetup{
pdftitle={A template for the arxiv style},
pdfsubject={q-bio.NC, q-bio.QM},
pdfauthor={David S.~Hippocampus, Elias D.~Striatum},
pdfkeywords={First keyword, Second keyword, More},
}

\begin{document}
\maketitle
\begin{abstract}
	Efficient management of end-of-life (EoL) products is critical for advancing circularity in supply chains, particularly within the construction industry where EoL strategies are hindered by heterogenous lifecycle data and data silos. Current tools like Environmental Product Declarations (EPDs) and Digital Product Passports (DPPs) are limited by their dependency on seamless data integration and interoperability which remain significant challenges. To address these, we present the Circular Construction Product Ontology (CCPO), an applied framework designed to overcome semantic and data heterogeneity challenges in EoL decision-making for construction products. CCPO standardises vocabulary and facilitates data integration across supply chain stakeholders enabling lifecycle assessments (LCA) and robust decision-making. By aggregating disparate data into a unified product provenance, CCPO enables automated EoL recommendations through customisable SWRL rules aligned with European standards and stakeholder-specific circularity SLAs, demonstrating its scalability and integration capabilities. The adopted circular product scenario depicts CCPO's application while competency question evaluations show its superior performance in generating accurate EoL suggestions highlighting its potential to greatly improve decision-making in circular supply chains and its applicability in real-world construction environments.
\end{abstract}


\section{Introduction} \label{intro}

 End-of-life (EoL) management of construction products is becoming more crucial as the sector plays a major role in global pollution, responsible for 36\% of final energy consumption and 39\% of energy-related CO\textsubscript{2} emissions \cite{iea}. Attaining circular economy (CE) objectives in this sector requires EoL decision-making, which is defined as the assessment and execution of actions to extend a product’s lifecycle \cite{petri_nets}. Circular supply chains entail restorative processes like refurbishment, reuse and recycling \cite{Ellen_mac} and require robust EoL strategies to guarantee sustainability and economic feasibility. Essential factors for effective EoL decisions consist of product provenance, stakeholder collaboration, effective data management and standardization, and compliance with regulations \cite{Saman_traceability, konie_circular, Env_impacts, wang_rfid}. 
 
 \begin{figure}[h]
  \centering
  \includegraphics[scale=0.5]{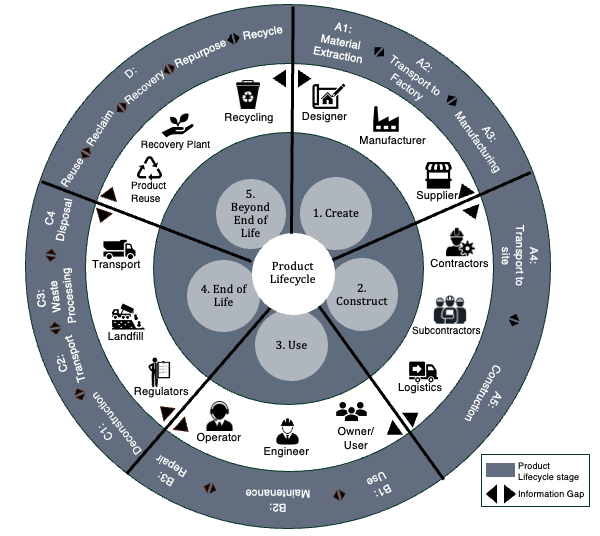}
  \caption{Product Lifecycle Phases}
  \label{fig:plc}
\end{figure}

Effective EoL management is a complex, "fuzzy" process \cite{EoL_1,decision_CE} consisting of significant challenges described below: 
\\[0.1cm] 
\textbf{1. Information exchange.} EoL decisions require comprehensive data collection across a product's lifecycle. However, this is hampered by incomplete information, fragmented data systems, and interoperability problems. Figure \ref{fig:plc} depicts the various stages across a product lifecycle (PLC) and the sources of information. For informed EoL decision-making, information must be exchanged seamlessly. Still, in practice, that is not the case and the information is often siloed, impeding accurate EoL decision-making.
\\[0.1cm]
\textbf{2. Lack of standardisation.} The lack of universally accepted standardised terminologies in construction leads to inconsistencies in data representation. This hinders effective data integration leading to challenges in applying consistent EoL decision-making frameworks \cite{eol_building} and effectively, hindering circularity efforts.
\\[0.1cm]
\textbf{3. Heterogenous product data.} EoL decision-making requires both qualitative (e.g. product health) and quantitative (e.g. emissions) data with either static or dynamic data flows \cite{decision_CE, petri_nets}. For example, product composition data is static whereas fluctuating market demand is dynamic. Processing this diverse data set presents difficulties, requiring sophisticated automation systems.

Although technologies like IoT and blockchain enable real-time data sharing and traceability, difficulties in semantic and protocol interoperability persist \cite{industry_4.0, provTracing, IoT_protocols}.  New information exchange solutions like Digital Product Passports (DPP), Environmental Product Declarations (EPD), and Material Passports (MP) are still hindered by standardisation issues and supply chain data silos \cite{sultan_data}.

This paper presents the \textit{Circular Construction Product Ontology (CCPO)} as a step towards addressing these challenges. Ontologies are a formal representation of domain knowledge and through standardised definition of terms and their inter-relationships, they provide contextual knowledge for informed decision making \cite{what_is_ontology, ontology_Knowledge}. CCPO is a framework for organizing construction product data, enabling information exchange among supply chain stakeholders and automated EoL decision-making via modifiable rules. While ontological models have been applied in construction for decision-making in various ways, this research proposes a unique approach to automated EoL decision-making leveraging ontology's strengths with formulated rules based on circularity goals. The contributions of this paper include:
\begin{itemize}[wide]
  \item A novel ontology that extends existing frameworks with rule-based reasoning capabilities and data aggregation across supply chain stakeholders. This facilitates semantic interoperability and enables informed decision-making in circular supply chains.
  \item A model that enhances product provenance by adapting existing ontologies to represent all standardised stages in a construction circular supply chain with consolidated PLC information across multiple stakeholders.
  \item A rule-based framework for automated EoL decision-making aligned with circularity standards and adaptable to varying stakeholder service level agreements (SLAs) for enhanced scalability and applicability in real-world construction scenarios.
  \item A scenario-based development methodology validated through expert review and real-world use cases, showcasing CCPO's practical relevance and effectiveness in addressing complex EoL decision-making challenges in construction.
\end{itemize}

The remainder of this paper is as follows: Section \ref{relatedWork} discusses related work on ontologies in circular supply chains. Section \ref{background} provides relevant background literature. Section \ref{implementation} details the implementation approach while Section \ref{evaluation} evaluates the developed model.  Finally, Section \ref{conclusion} concludes with limitations and future work.

\section{Related Work} \label{relatedWork}

Limited research was found that directly apply semantic technologies, enhanced with SWRL rules for EoL handling in construction circular supply chains. However, several existing decision-making ontology models provided a basis for the CCPO development. Ontology can improve traditional decision-support systems by providing a rich semantic representation of domain knowledge \cite{web_onlology, Tejal_1, Tejal_2}. Formal ontologies use standards such as the Web Ontology Language (OWL) and Resource Description Framework (RDF) to represent complex knowledge in the form of concepts (entities) and their dynamic relations (properties) \cite{onthology_101}. The current version of OWL is OWL 2 and OWL 2 DL is the decidable fragment of OWL 2 \cite{owl_2, SROIQ,ontology_engineering}. It enables object-oriented modelling with consistency checking and support for complex properties like symmetry, asymmetry, and cardinality \cite{DL_lite}. Furthermore, OWL 2 can be extended with Semantic Web Rule Language (SWRL) to define extensible rules to model complex scenarios, explicate further implicit knowledge, and enable decision-making through automated reasoning using reasoners like Pellet and Hermit \cite{swrl}.

The Circular Economy Ontology Network (CEON) \cite{coreOntologies} provides a foundation for cross-industry data exchange. While its innovative approach applies to various industries, its generality limits its direct usefulness in EoL processes. CCPO builds upon CEON to offer a more targeted solution for EoL handling. The ontology framework in \cite{Ontology_CE} addresses EoL processing in supply chains. However, this work only monitors CE progress rather than activities. Also, the framework requires further work in addressing terminology conflicts. The ontology model in \cite{Ontology_AHP} adopts the Analytic Hierarchy Process (AHP) for a structured decision-making process in the reuse and recycling of construction machinery parts. One weakness of this model is that it only focuses on assessing the quality of decisions, disregarding supply chain economic and environmental implications. Similarly, the ontological model in \cite{disassembly} uses partial destructive rules to optimise the disassembly of automobile traction batteries (ATB). However, challenges persist such as knowledge base restrictions which impact their system's efficacy. Ontology-based models can also be combined with various technologies for enhanced functionality. For example, BiOnto \cite{BiOnto} enables collaboration between scientists, enterprises and policymakers. It uses the Aspose and Text Razor APIs for text extraction and semantic analysis respectively. However, BiOnto only focuses on bioeconomy without addressing circularity concerns. The PLM ontology \cite{ontology_building} also employs OWL and SPARQL for query processing. Though the model promotes sustainable manufacturing through enhanced data sharing, it's application in circularity practices is limited.  The ontology and blockchain framework in \cite{ontology_BC} enable secure product tracking in supply chains. However, the researchers reported challenges in converting ontology concepts into smart contracts. The EAGLET ontology \cite{ontology_supply} combines RFID and IoT for real-time product tracking while \cite{ontology_Knowledge} applied ontology for sustainable supplier selection via natural language processing (NLP) techniques.

Reusing upper ontologies boosts semantic clarity, consistency and development efficiency. The Basic Formal Ontology (BFO) \cite{bfo} and Common Core Ontologies (CCO) \cite{cco} are scalable models reusable across various applications. BFO categorises real-world items as either occurring in time (occurents) or persisting over time (continuants) and CCO uses twelve mid-level ontologies covering artefacts, agents, events, geography, and time. CCPO extends these ontologies including the provenance (PROV) \cite{prov} and building product ontology (BPO) \cite{bpo} for data aggregation and product tracking capabilities to support accurate EoL handling aligned with industry standards such as EN15804 \cite{EN_15804} and EC PEF  \cite{EC_PEF}. This positions CCPO as a critical advancement in ontology-based frameworks and a practical tool for applied computing in construction.

\section{End-of-Life Decision-making and Relevant Standards in Construction Supply Chains} \label{background}
Current supply chain EoL approaches such as eco-efficiency, lifecycle assessment (LCA) and multicriteria decision analysis (MCDA) \cite{decision_CE} are often limited by manual data aggregation and lack of interoperability between systems. Eco-efficiency seeks to minimise environmental and economic impact while maintaining product value \cite{Biomass_conversion}. The LCA approaches offer comprehensive evaluations across a product's lifecycle by collecting scientific data. MCDA schemes integrate both quantifiable and non-quantifiable factors for sustainability, assisting decision-makers in navigating intricate EoL scenarios. Other approaches include adopting modular construction designs and material passports. Despite the existing approaches, EoL handling processes remain largely manual requiring data aggregation and analysis across diverse sources. The CCPO addresses this challenge through automated decision flows and standardised semantic exchanges with digital tools like DPPs and EPDs for effective EoL processing. The choice of the EoL approach remains product-specific hence endeavours in promoting circularity must be iterative and progressive rather than a one-time approach. As such, this work focuses on reuse and recycling options with a blend of eco-efficiency and LCA approaches for a rule-based framework. Product, market and process variables are expedient for effective EoL handling \cite{petri_nets} and include \textit{ProductHealthState} (\textit{$P_{H}$}), \textit{Recyclability}, \textit{reference service life} (\textit{RSL}) and \textit{OperatingDuration} (\textit{OD}). As products reach the EoL, their health (\textit{$P_{H}$}) decreases as modelled in Equation \ref{eq1}. \textit{$P_{H0}$} denotes the initial health and $\alpha$ is the rate of health decline. Recyclability refers to a product's capability to be recycled and \textit{RSL} is the expected product use duration before needing repair or replacement. The \textit{OD} or \textit{Actual Service Life (ASL)} measures the product's use period in years. Economic feasibility and market demand are also key factors. For example, high \textit{MarketDemand} for recycled products gives recycling a higher priority over other EoL options. Process variables such as \textit{EnvironmentalImpact} and availability of recycling options further influence EoL decisions. 

\begin{equation} \label{eq1}
    \textit{$P_{H}$} = \textit{$P_{H0}$} * e^{-a.OD} 
\end{equation}

\begin{figure}[h]
  \centering
  \includegraphics[scale=0.5]{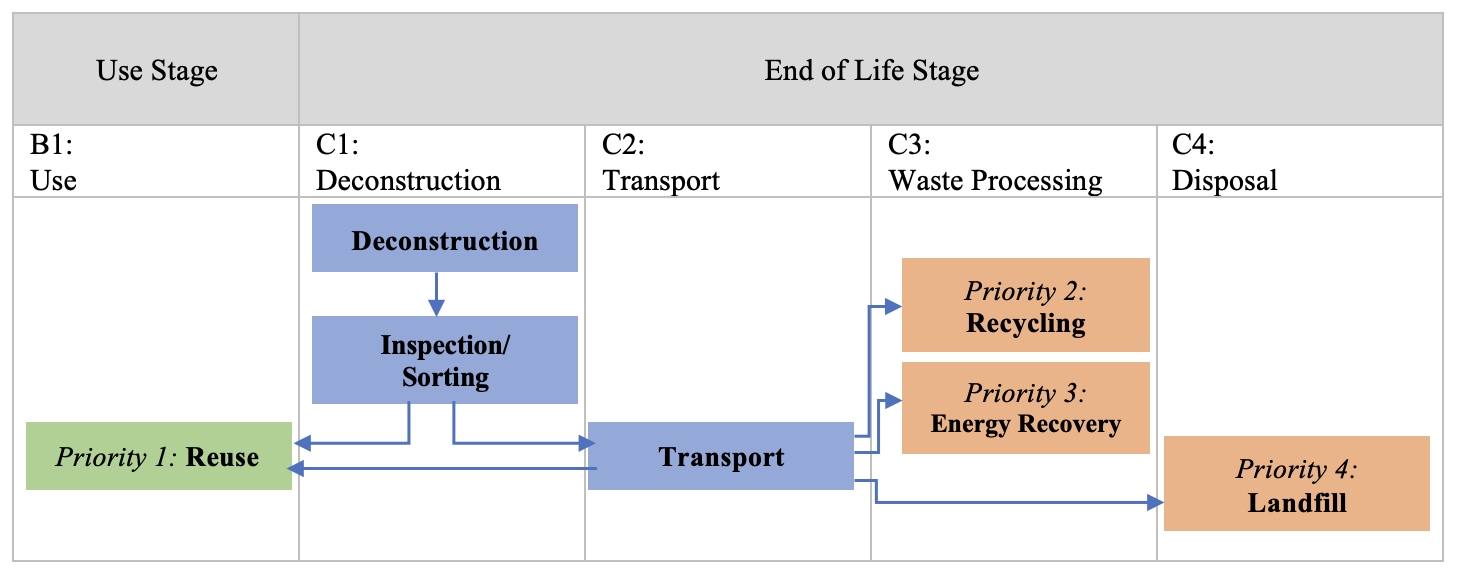}
  \caption{EoL Handling Priority Hierarchy}
  \label{fig:eol_priority}
\end{figure}

The rule-based CCPO model integrates key industry standards like the  European Commission's Product Environmental Footprint (EC PEF)  \cite{EC_PEF} and the EN 15804 \cite{EN_15804} to ensure compliance with established environmental metrics. The EN 15804 offers a structured LCA framework for construction products encompassing various stages (Modules A-D) of the PLC and promoting EPDs. The 2019 version of this standard broadened its scope to include 19 impact categories for a thorough perspective of environmental impact with module C targetting EoL considerations. EC PEF provides a comprehensive method for standardisation applicable to various products. It provides the Carbon Footprint Formula (CFF) to compare emissions and impacts. This formula includes metrics like  \textit{$E_{recycling, EOL}$}, \textit{$E_{D}$} and \textit{Q} for emissions from recycling at EoL, emissions from disposal and product quality respectively. The European Union's waste hierarchy \cite{waste_hierarchy} prioritises prevention, reuse and recycling over energy recovery and disposal processes. Figure \ref{fig:eol_priority} depicts how this hierarchy aligns with EN 15804 to inform EoL strategies. Furthermore, CCPO's informed decision-making aligns with the recent Ecodesign for Sustainable Products Regulation (ESPR) \cite{espr} for promoting circularity, limiting waste generation and improving information availability. Integrating these standards into the CCPO addresses the practical challenges of EoL decision-making in construction by providing a standardised, automated assessment framework. This approach fosters transparency and trust among stakeholders, and bridges the gap between data-driven decision-making and real-world implementation in sustainable construction practices.

\section{Implementation} \label{implementation}

\begin{figure}[t]
  \centering
  \includegraphics[scale=0.7, trim=9.8cm 5.5cm 8.2cm 3cm, clip]{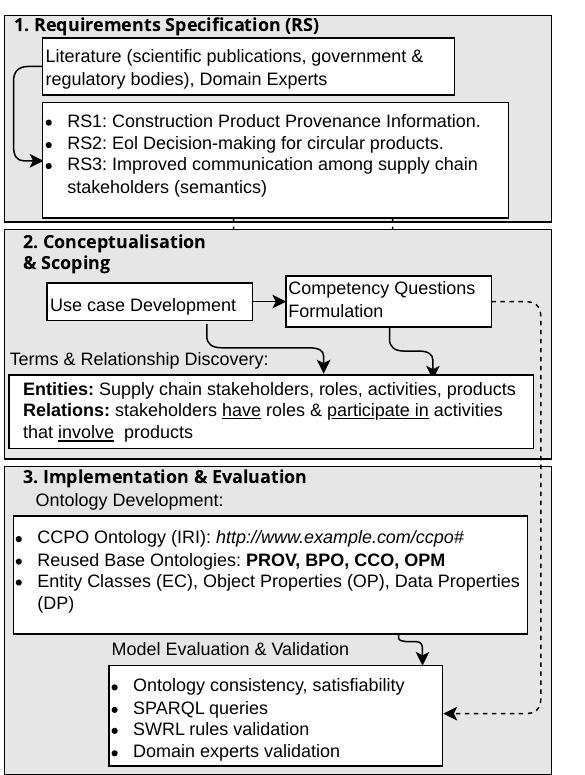}
  \caption{Development Approach}
  \vspace{-4ex}
  \label{fig:dev_framework}
\end{figure}

The CCPO was developed based on the METHONTOLOGY \cite{methont} and NeOn \cite{NeOn} methodologies. METHONTOLOGY emphasizes the reuse of existing ontologies whereas NeOn follows a modular approach driven by competency questions. The development approach in Figure \ref{fig:dev_framework} outlines the key steps in the CCPO implementation. 
\\[0.1cm]
\textbf{Specification:} An initial domain analysis was carried out involving literature and online resources like the Ellen MacArthur Foundation \cite{Ellen_mac}, CIPS \cite{glossary} and the construction wiki \cite{iea}. Three domain experts- a building engineer, a supply chain business expert and staff from a steel manufacturing company were also interviewed. The identified gaps in construction circular supply chains influenced the research specifications or goals: \textbf{RS1:} The need for effective technology solutions for full product lifecycle overview. \textbf{RS2:} Models to support EoL decision-making. \textbf{RS3:} Improving information exchange with an emphasis on semantics.
\\[0.1cm]
\textbf{Conceptualisation \& Scoping:} After the domain analysis, a scenario based approach was developed which highlights the necessity of gathering PLC information for accurate EoL decision-making. The Insulated Wall Panel (IWP) scenario presented in subsection \ref{scenario} also provided a reference for discovering the ontology's relevant terms and relationships. The competency questions (CQs) below were created from the research specifications to target various preferred capabilities of the CCPO and ultimately, provide a scope for its development. These CQs were formalized into SPARQL queries and SWRL rules to assess the CCPO's operation.

\begin{itemize}
    \item \textbf{CQ1:} What are the constituent virgin and non-virgin materials in the IWP?
    \item \textbf{CQ2:} What is the up-to-date provenance information for the IWP?
    \item \textbf{CQ3:} What are the available information-bearing artifacts associated with the IWP in its current state?
    \item \textbf{CQ4:} What is the current health state of the IWP and the market demand for its recycled product?
    \item \textbf{CQ5:} Is there an existing Manufacturers’ EoL strategy for the IWP?
    \item \textbf{CQ6:} Can the developed model suggest the end-of-life handling strategy for the IWP which is at its end-of-use stage?
\end{itemize}

\begin{figure}[t!]
  \centering
  \includegraphics[scale=0.55, trim=6.3cm 7.45cm 7.2cm 5.6cm, clip]{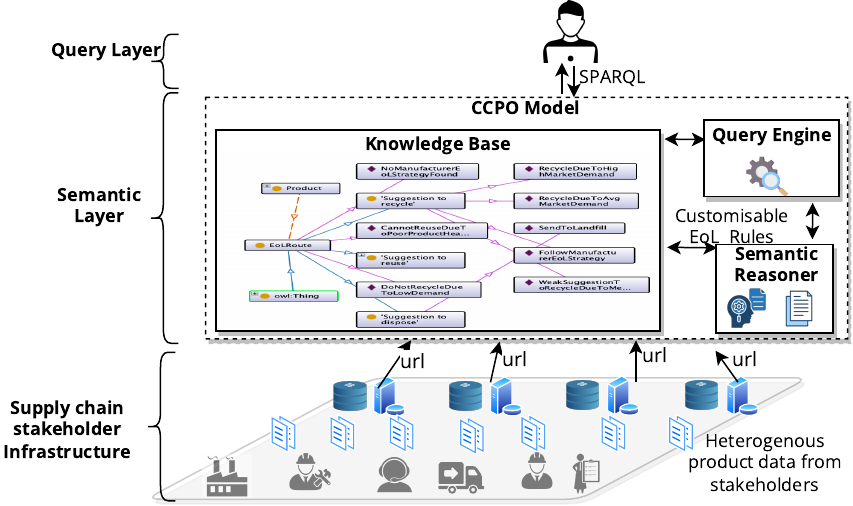}
  \caption{Integrated EoL Decision-making Framework}
  \label{fig:integratedFramework}
\end{figure}

\textbf{Implementation, Integration  \& Evaluation:} The CCPO was implemented in OWL using Prot\'eg\'e \cite{protege} and incorporating existing ontologies like PROV, BPO, CCO, Digital Construction Building Material Ontology (DICBM) \cite{dicbm} and the Ontology for Property Management (OPM) \cite{opm}. Classes, object properties and data properties were defined to represent the domain terminologies, relations and characteristics while SPARQL queries were developed to query circularity and EoL metrics. CCPO's decision-making framework, presented in Figure \ref{fig:integratedFramework}, comprises three key layers; query, semantic and infrastructure. In the query layer, users access the CCPO model via SPARQL queries which enable precise information retrieval through pattern matching. These queries can extract both schema-level metadata and instance data, such as product composition and provenance. Advanced SPARQL queries like filtering, aggregation and optional patterns enable queries versatility for complex data retrieval tasks. The semantic layer forms the core of the CCPO, capturing relevant entities like actors, products and locations. It also models hierarchies, object and data properties essential for EoL decision-making. The query engine processes SPARQL queries while the semantic reasoner makes inferences and automatically suggests EoL routes based on predefined SWRL rules aligning with stakeholder SLAs and circularity standards. The infrastructure layer comprises cloud storage systems managed by supply chain stakeholders. The innovative CCPO model integrates these layers to automatically provide EoL recommendations for construction products, thereby ensuring compliance with organisational SLAs and circularity standards. Though the CCPO is unable to extract product documents directly from these systems, it achieves provenance and data aggregation via its instance data and providing links to product data. This ensures that data remains under the owner's control. This streamlines data management by eliminating the need for data duplication among stakeholders and ensuring secure and controlled access. Synthesized data which aligns with the real-world use case was used for the ontology validation against the CQs. The validation of the CCPO model further assesses the ontology's quality, usability and fitness for the intended purpose. The Pellet reasoner \cite{pellet} was used for ontology consistency checking and deriving the solutions for the SPARQL queries. The final model was also validated by having domain experts review it against how adequately it addresses the research specifications. 

\subsection{The Insulated Wall Panel Scenario}
\label{scenario}

\begin{figure*}[ht!]
  \centering 
  \includegraphics[scale=0.55, trim=4cm 0.95cm 2cm 11cm, clip]{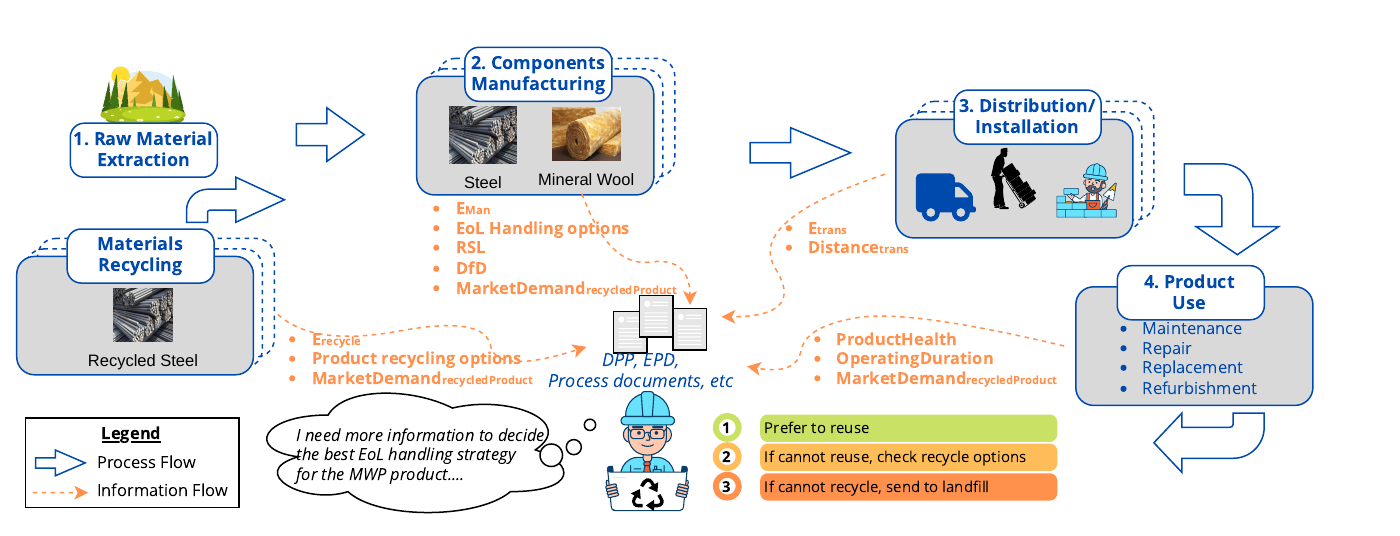} 
  \caption{IWP Scenario}
  \vspace{-3ex}
  \label{fig:iwp_scenario} 
\end{figure*}

The Insulated Wall Panel (IWP) scenario demonstrates the intricacies of EoL decisions for construction products, particularly in collecting heterogeneous product data from various PLC stages. The scenario is depicted in Figure \ref{fig:iwp_scenario} and includes the various data sources, process flows, stakeholders and decision points. The IWP is a prefabricated wall system with two steel facings and a mineral wool core. It was selected for this scenario because of its circularity design- its components are easily replaceable, reusable, or recyclable. The scenario was developed in partnership with construction specialists on an EPSRC-funded project. 

A building engineer of Company X is assigned to identify the most efficient EoL strategy for the IWP product. To align with the company's sustainability goals, the strategy must prioritise product reuse and recycling over disposal. This requires comprehensive data collection and analysis from various stages of the supply chain.
Data recording starts at the manufacturing phase capturing metrics such as material composition, design for disassembly, recyclability and Global Warming Potential (GWP). These data points may be extracted from existing tools like the DPP and EPD. The engineer also evaluates market demand for the product and its reference service life (RSL) to determine the feasibility of recycling. Transport emissions data from logistic providers aid the decision-making process by comparing the environmental impact of different EOL routes. At the product use stage, maintenance data, IoT sensor recordings and visual inspections are used to assess the product's condition and total operating duration or actual service life (ASL). This information is essential to determine whether the product can be reused or needs other EoL route considerations. Subsequently, the environmental impacts like emissions from the selected EoL route may be evaluated from estimated or historic data.

The decision-making process requires integrating different data types. This includes dynamic, non-quantitative metrics like product health which changes over time and requires appropriate analytical schemes. Carbon emissions are however static and qualitative. Also, market demand for recycled products often relies on estimations derived from historical data, adding more complexity to the decision-making process. One major challenge arises from the decentralisation of information across multiple stakeholders resulting in data losses at various PLC stages. This fragmented data hinders engineers from accessing complete information which leads to making assumptions in their EoL decisions. The scenario highlights the need for a standardised ontology framework that guarantees effective information exchange and semantic interoperability compliant with the principles of circular economy. Various terms used in this scenario and in the implementation were adapted from authoritative sources such as the Waterman group \cite{waterman} and others in \cite{Ellen_mac,glossary,design_building,eco_portal} to establish a shared vocabulary for supply chain stakeholders. Some pertinent ones have been presented here: 
\\[0.1cm]
  \textbf{Design for Disassembly (DfD)} – specifies any provision by manufacturers on how an assembled product may be disassembled.
  \\
  \textbf{Material} – refers to any raw or processed substance used in the construction of buildings or other structures to form elements.
  \\
  \textbf{Product} – refers to a tangible object sold by a manufacturer or supplier to be used in the construction of buildings or other structures.
  \\
  \textbf{Grouped Component} - refers to a product comprising of multiple components and fulfils its function as a grouped component.
  \\
  \textbf{Component} - represents the description of an object that is part of a Grouped Component or the product itself. Whenever such an object is modelled, it should ideally be given one of the subclasses of Product to further specialise the object's characteristics.

\begin{figure*} 
  \centering 
  \includegraphics[scale=0.67,  trim=1.6cm 1.2cm 4cm 13.5cm, clip]{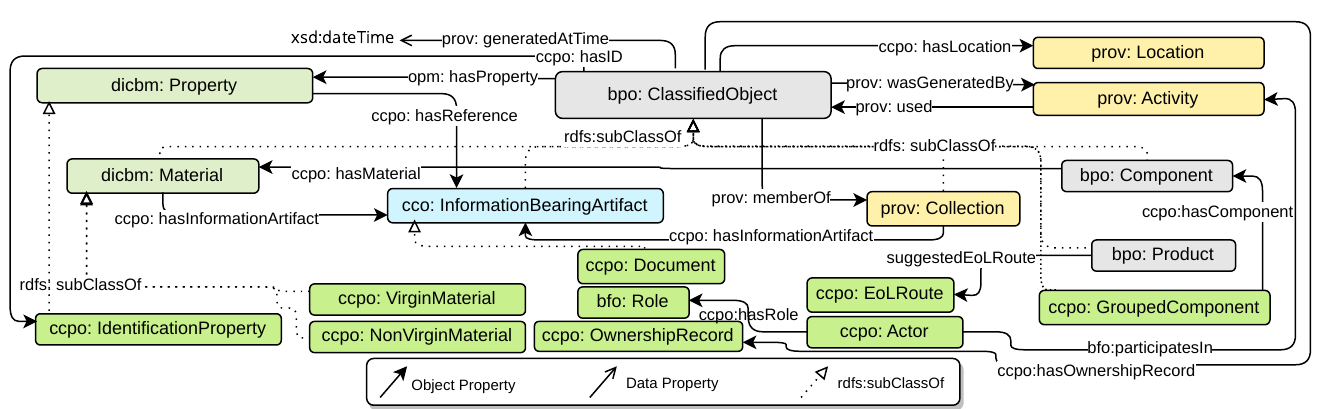} 
  \caption{The CCPO Ontology Main Classes} 
  \label{fig:CCPO_mainClasses}
\end{figure*}

\subsection{The Construction Circular Product Ontology (CCPO)}

The CCPO framework automates EoL decision-making within decentralised and heterogeneous information networks by implemented SWRL rules and PLC data integration. This automation reduces reliance on manual data handling, enhances decision accuracy and ensures alignment with sustainability goals, demonstrating CCPO's practical application in complex construction scenarios.

Figure \ref{fig:CCPO_mainClasses} shows CCPO's key classes and relations for decision-making with new CCPO definitions and imported classes distinguished by prefixes and colour-coded. Ontology merging ensures the extensibility and consistency of the CCPO so it may be queried in combination with other ontologies such as the building topology ontology \cite{bot}. CCPO aligns with BFO such that $\textit{ccpo:Actor} \sqsubseteq \textit{bfo:IndependentContinuant}$ and $\textit{prov:Activity} \sqsubseteq \textit{bfo:OccurrentProcess}$. All top classes are set as disjoint to avoid ontology inconsistencies. \textit{Products} are classified as either \textit{Components} or \textit{GroupedComponents} for a single component or an assembly of multiple components: $\textit{GroupedComponent} \sqsubseteq \textit{Product} \sqcap (\geq 2 \ \textit{hasComponent}.\textit{Component})$. Product provenance is captured as a chain of products and generating activities where activities \textit{prov:used} materials or products to generate new products: $\textit{Product} \sqsubseteq \exists \textit{wasGeneratedBy}.\textit{Activity}$ and $\textit{Activity} \sqsubseteq \exists \textit{used}.(\textit{Product} \sqcup \textit{Material})$. The object property, $\textit{ccpo:wasInvolvedInActivity}$  represents relations between products and activities which do not modify the product or create new products. Provenance is also captured by a trail of product ownership changes using the $\textit{hasOwnershipRecord}$ property. The \textit{OwnershipRecord} and \textit{Activity} classes are finite-time processes and hence captured using the \textit{prov:startedAtTime} and \textit{prov:endedAtTime} data properties. The CCPO also links products and materials with their related product data- \textit{cco:InformationBearingArtifacts} (IBA) which may be a \textit{Document} or \textit{Barcode}. The CCPO uses URL links to aggregate dispersed product data from various stakeholders involved in the product's lifecycle to support full provenance: $(\textit{Product} \sqcup \textit{Material}) \sqsubseteq \exists \textit{hasInformationArtifact}.\textit{InformationBearingArtifact}$. Documents are connected to location sources such as a URL by the \textit{ccpo:hasLocation} relation. It is assumed that supply chain actors implement appropriate data access mechanisms for users of CCPO. The \textit{dicbm:Property} class captures product properties. The EoL decision-making is automatically inferred from SWRL rules acting on available product data. This is discussed in Section \ref{algorithm}.

\subsection{Rule-based EoL Decision-making} \label{algorithm}

Algorithm \ref{eol_alg} leverages CCPO's rule-based reasoning to evaluate key environmental and economic metrics like carbon emissions and market demand for recycled products. By prioritising reuse, the algorithm ensures EoL decisions are aligned with sustainability goals demonstrating CCPO's practical utility. The algorithm initially considers base requirements for EoL handling but can be customised to incorporate additional SWRL rules allowing for tailored decision-making that meets specific stakeholder needs.

The algorithm first evaluates the $P_H$, $ASL$ and $RSL$ at the end-of-use stage. For products reaching their $RSL$, $If$ $P_H == "green"$, then a strong recommendation for reuse is given. A $weak\_reuse$ recommendation occurs when $ASL >= RSL$ and $P_H == "amber"$. This suggests the product may need some maintenance or refurbishment soon. If prior conditions fail, the product cannot be reused. Values for $P_H$ align with the Royal Institution of Chartered Surveyors' (RICS) \cite{rics} classification of the state of buildings and building components. A condition rating of 1 ($green$) means no repair is needed and the product can be maintained following normal practices. A condition rating of 2 ($amber$) signifies the product may need repair or replacement soon while a rating of 3 ($red$) is used for products with serious defects that need urgent investigation.

If reuse checks fail, the ontology checks the availability of the manufacturer's EoL handling specifications. $If$ $strategy\_found = true$, the model returns the location of the related document.
If no such specification is found, the ontology evaluates the economic and environmental conditions for either recycling or landfill. $If$ $market\_demand == "high" $ for the recycled material/ product and the product is designed with circularity in mind, ($DfD == "true"$), then the algorithm suggests product recycling otherwise the suggestion to dispose of the product is given.

\begin{algorithm}
\SetAlgoLined
\footnotesize
\caption{End-of-Life Decision Making}
\label{eol_alg}

\SetKwFunction{FEvaluateHealth}{EvaluateHealth}
\SetKwFunction{FCheckEoLStrategy}{CheckEoLStrategy}
\SetKwFunction{FMarketDemand}{MarketDemand}
\SetKwFunction{FCompareEmissions}{CompareEmissions}
\SetKwFunction{FEoLDecision}{EoLDecision}

\SetKwProg{Fn}{Function}{}{}

\Fn{\FEvaluateHealth{product}}{
    \uIf{product.$P_H$ == "green" $\land$ product.$ASL$ < product.$RSL$}{
        \KwRet{reuse\_strong}\;
    }\uElseIf{product.$P_H$ == "amber" $\lor$ product.$ASL$ >= product.$RSL$}{
        \KwRet{reuse\_weak}\;
    }\Else{
        \KwRet{reuse\_not}\;
    }
}

\Fn{\FCheckEoLStrategy{product}}{
    \uIf{product.EoLStrategy.exists}{
        \KwRet{strategy\_found}\;
    }\Else{
        \KwRet{strategy\_not\_found}\;
    }
}

\Fn{\FEoLDecision{product}}{
    \textit{health\_decision} $\leftarrow$ \FEvaluateHealth{product}\;
    \uIf{health\_decision == "reuse\_strong"}{
        \KwRet{"Reuse: Strong"}\;
    }\uElseIf{health\_decision == "reuse\_weak"}{
        \KwRet{"Reuse: Consider refurbish"}\;
    }\uElseIf{\FCheckEoLStrategy{product} == "strategy\_found"}{
        \KwRet{"EoLStrategy Doc Location: URL"}\;
    }\uElseIf{market\_demand =="high" $\lor$ market\_demand="avg" $\land$ DfD == "True"}{
        \KwRet{"Recycle"}\;
    }\Else{
        \KwRet{"Landfill"}\;
    
    }
    }
\end{algorithm}

\section{Evaluation} \label{evaluation}

The CCPO model was evaluated on various parameters to ensure its accuracy, effectiveness and usability. This involved assessing the model's structure, expressivity, reasoning capabilities, interoperability and scalability. Subsequently, a task-based evaluation \cite{ont_eval, ont_prin} tested the ontology's competency in completing specific tasks thus highlighting its performance in EoL decision-making.


Table \ref{tab:struct_metrics} summarizes the CCPO's structural metrics. Their impact on semantic interoperability, data integration and performance are discussed in this section. Consistency checks using the Pellet reasoner verified the model's logical soundness, reliability and interoperability with other ontologies. The CCPO's 88 classes and 35 object properties comprehensively model real-world entities and interactions in the circular supply chain domain.  This extensive class structure enhances the model's reliability and accuracy of EoL recommendations in construction environments. A data property count of 27 ensures rich descriptions of entities for enhanced ontology versatility and applicability to real-world data systems. Despite the moderate individual count of 38, it is sufficient for a comprehensive validation, thereby guaranteeing logical consistency. The high axiom count of 781 reflects the complexity and maturity of this ontology. A lower count may be better for lightweight applications that prioritize efficiency and simplicity. However, for the complex construction circular supply chain, this count is necessary to adequately capture detailed relationships and rules for accurate EoL decisions.

\begin{table}[h]
  \small
  \caption{CCPO Structural Measures}
  \label{tab:struct_metrics}
  \centering
  \begin{tabular}{lr}
    \toprule
    Measure&Value\\
    \midrule
    Consistency\textit{(Pellet)} & \checkmark\\
    Class count & 88\\
    Object+Data property count & 35+27\\
    Individual count & 38\\
    Axiom count & 781\\
    DL Expressivity & \(\mathcal{ALCHIQ(D)}\)\\
  \bottomrule
\end{tabular}
\end{table}

The CCPO's well-defined schema is verified by its DL expressivity of  ($\mathcal{ALCHIQ(D)}$). It denotes the model's support for basic attributive language ($\mathcal{AL}$), complex concepts ($\mathcal{C}$), role hierarchies ($\mathcal{H}$), inverse roles ($\mathcal{I}$), qualified number restrictions ($\mathcal{Q}$) and support for data properties  ($\mathcal{D}$). The CCPO ensures consistency via constraints. For example, a cardinality constraint ensures that all products with a minimum of 2 components are automatically classified as a \textit{GroupedComponent}. Inverse roles and role hierarchies enable bidirectional relationships for sophisticated reasoning. Although computationally intensive, this expressivity is essential for complex reasoning in this domain. The Pellet reasoner mitigates the computational challenges by ensuring efficient processing. Semantic interoperability is a key strength of the CCPO which is achieved through several strategies; The ontology utilizes standardised terminologies for effective semantic exchange between supply chain actors to support EoL management. Secondly, support for various data types allows the CCPO to map and convert external data into its schema. CCPO's detailed knowledge representation, combined with its use of URLs to distributed product data, supports integration of data from multiple sources and stakeholders. Additionally, reusing ontologies and consultations with domain experts ensures the model's compliance with industry standards and compatibility with other systems or ontologies. 

\begin{figure*}[h]
    \centering
    \begin{subfigure}[b]{0.49\textwidth}
        \centering
        \includegraphics[width=\linewidth]{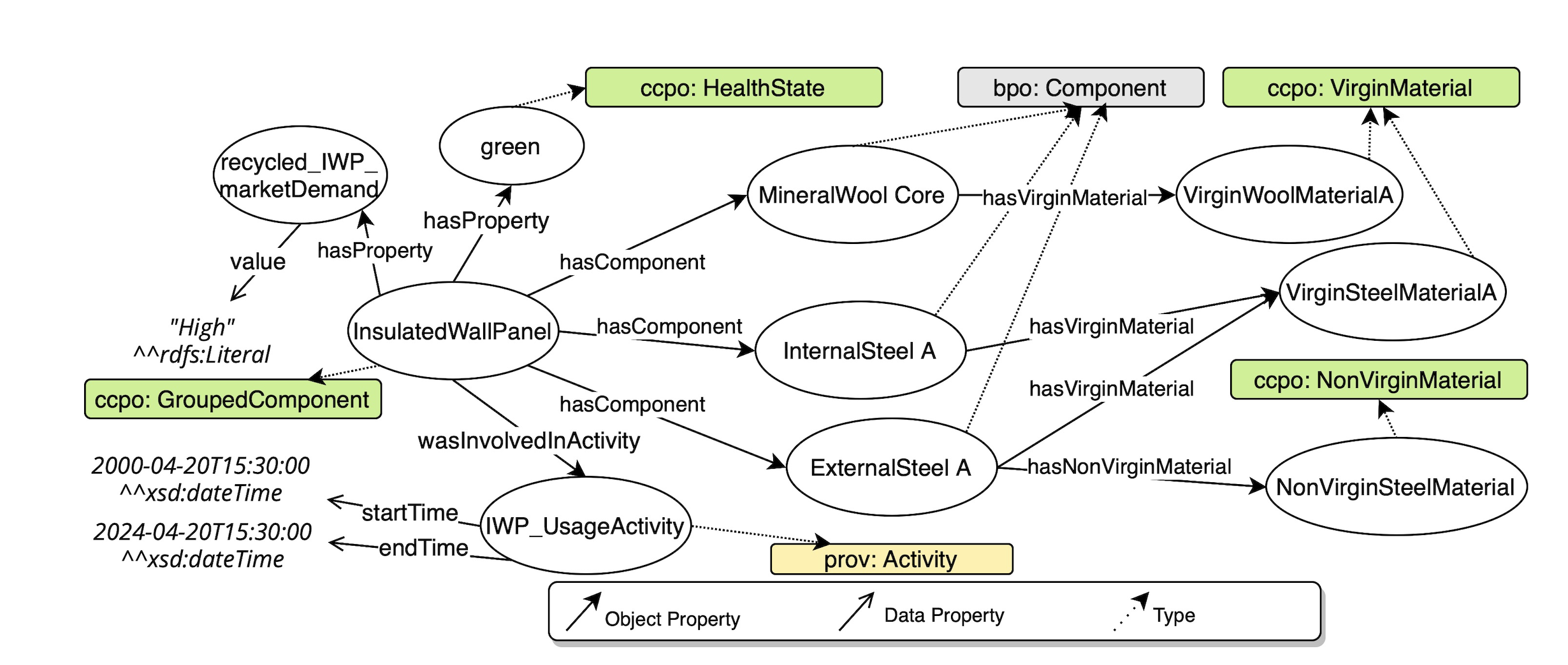} 
        \caption{CQs 1 and 4}
        \label{fig:CQa}
    \end{subfigure}
    \hfill 
    \begin{subfigure}[b]{0.49\textwidth}
        \centering
        \includegraphics[width=\linewidth]{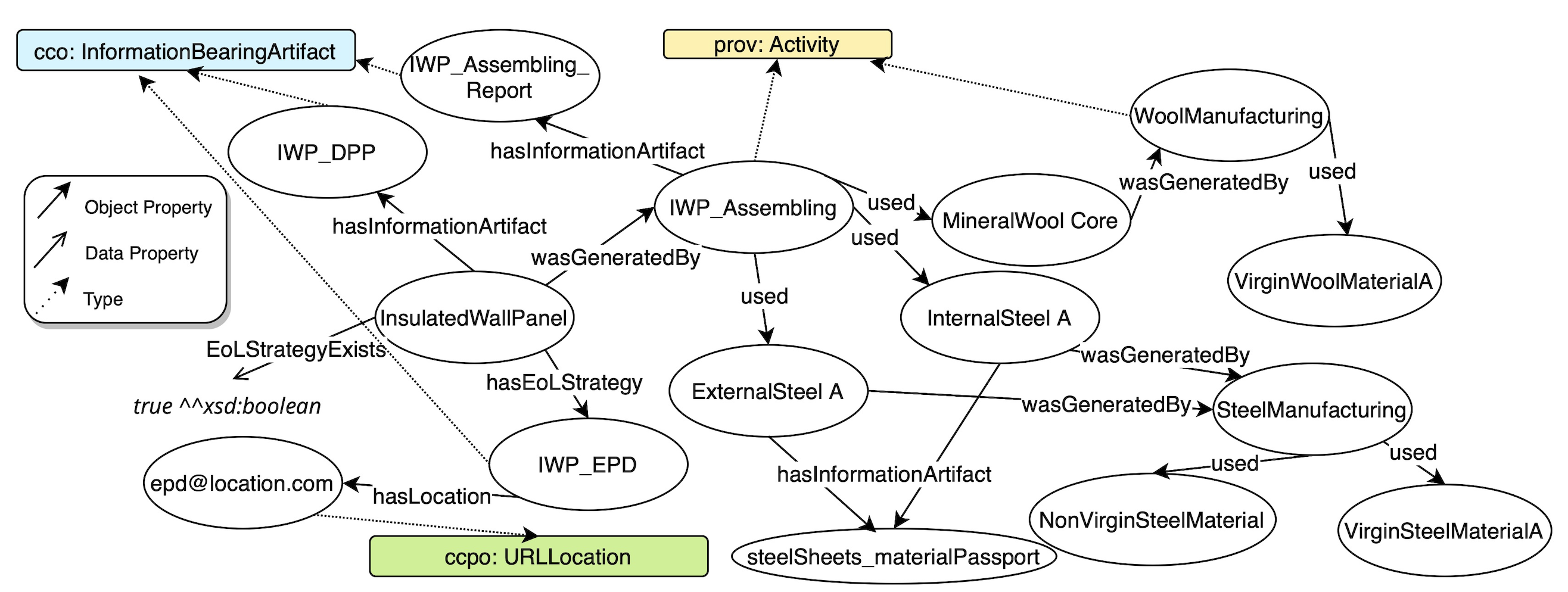} 
        \caption{CQs 2,3,5,6}
        \label{fig:CQb}
    \end{subfigure}
    \caption{CCPO Extract for CQs}
    \label{CQ}
\end{figure*}

\begin{figure*}[h]
    \centering
    \begin{subfigure}[b]{0.36\textwidth}
        \centering
        \includegraphics[width=\linewidth]{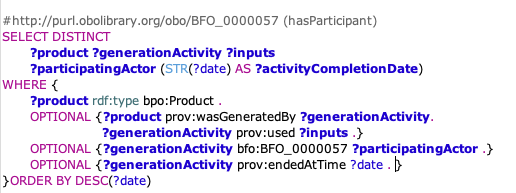} 
        \caption{SPARQL Query}
        \label{fig:CQquery}
    \end{subfigure}
    \hfill 
    \begin{subfigure}[b]{0.63\textwidth}
        \centering
        \includegraphics[width=\linewidth]{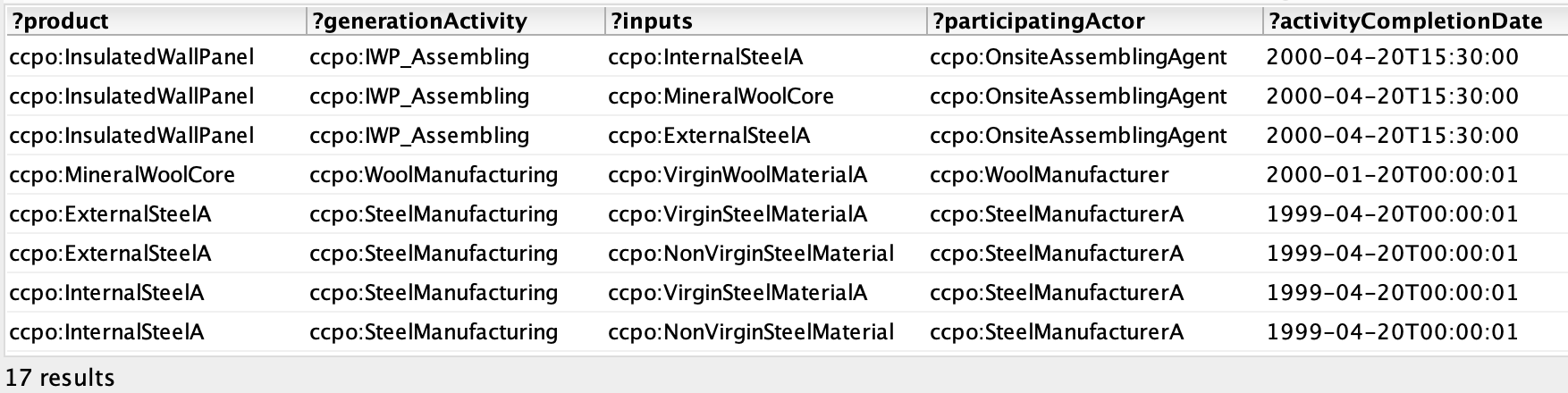} 
        \caption{Query results}
        \label{fig:CQresult}
    \end{subfigure}
    \caption{CQ2 SPARQL Query and Result}
    \label{task-based_evaluation}
\end{figure*}

\subsection{Task-Based Evaluation} \label{TB_eval}

The task-based evaluation of the CCPO model demonstrated its ability to meet project requirements while adhering to circularity principles of traceability, data connectivity and standards compliance. Furthermore, the flexibility of the CCPO allows it to be adapted to different construction projects demonstrating its potential for broader adoption in the industry. The CCPO was rigorously tested within the IWP scenario with the competency questions from Section \ref{implementation}, formulated into SPARQL queries as described below. However, the model is adaptable to other construction products.

\begin{figure*}[h]
  \centering 
  \includegraphics[scale=0.34]{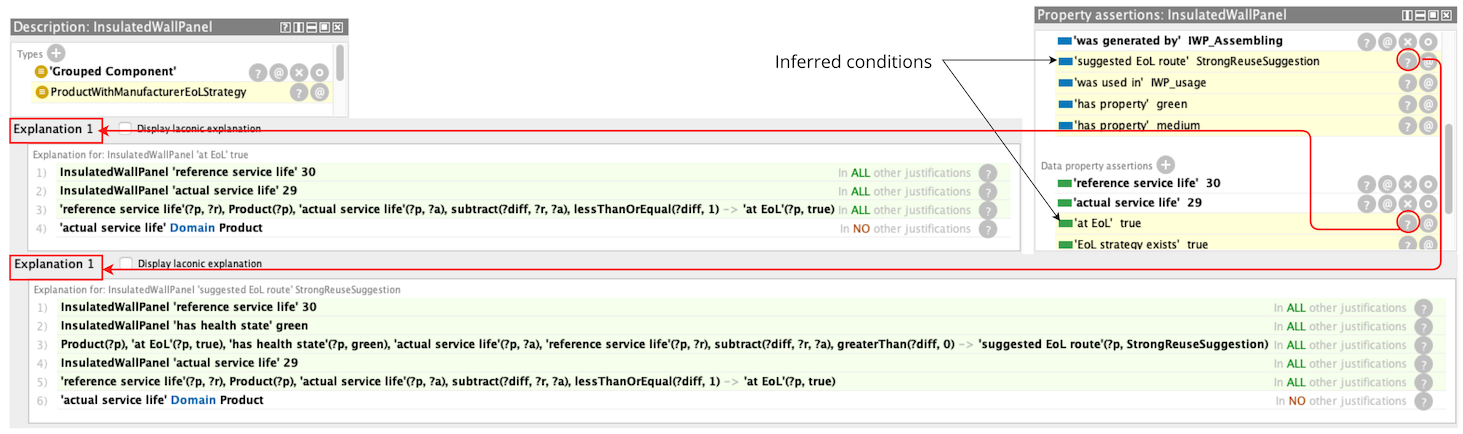} 
  \caption{Snapshot of the CCPO's Decision-making to Reuse a Product} 
  \label{fig:EoLdecision_making}
\end{figure*}

The Figures \ref{fig:CQa} and \ref{fig:CQb} depict the relevant CCPO individuals and relations used to answer the CQs. For instance, CQ1 uses the $\textit{ccpo:hasNonVirginMaterial}$ and $\textit{ccpo:hasVirginMaterial}$ properties to link components with their constituent materials for decision-making. The IWP grouped component uses the $\textit{ccpo:hasComponent}$ property to find its sub-components. CQ2 retrieves detailed PLC information demonstrating the CCPO's capability to collect and analyse complex provenance information. The SPARQL query in Figure \ref{fig:CQquery}  returns the result in Figure \ref{fig:CQresult} which lists associated generation activities in the PLC, input materials or products, associated stakeholders and activity times. The CQs were evaluated using complex SPARQL nested queries and optional patterns to overcome challenges with incomplete data which are typical in distributed systems. CQ3 utilises the \textit{ccpo:hasInformationArtifact} property to retrieve all IWP-related documents. This effectively establishes a connection between products and their documentation. The CQ4 queries the \textit{hasProperty} and more specialised \textit{hasHeathState} object relations to determine the product's current condition. 
The results from CQ5 and CQ6 are deduced using SWRL rules. The rules in table \ref{tab:swrl} demonstrate how SWRL rules support EoL decision-making by automatically deducing conclusions without explicit assertions. In Rule 1, the CCPO model automatically determines products reaching their EoL by comparing their ASL with the RSL. Products within one year of reaching their RSL are categorised as "at EoL". Subsequently, other SWRL rules act iteratively on the data, following the decision-making algorithm. OWL's  Open World Assumption denotes that if a fact is not known to be true, it is not automatically assumed as false. This presented a challenge in implementing the recycling options. We overcame this challenge by setting an intermediate classification to specify a product in bad health as \textit{"CannotReuse"}. The options for recycling are therefore only evaluated against products classified as such. Similarly, the rule for selecting the landfill option is evaluated only for products classified as \textit{"DoNotRecycle"}. Formal rules improves the CCPO model's expressivity in 2 ways. Firstly, conditions which cannot be captured using OWL may be inferred from rules. Rule 1 demonstrates this in finding products at their EoL. Secondly, rules add actionable knowledge to the ontology. By defining unambiguous criteria for EoL classification, the CCPO model supports accurate EoL management of construction products as shown in Figure \ref{fig:EoLdecision_making}.

The assessment of the CCPO using a series of competency questions showcased its efficacy and pertinence in delivering precise end-of-life recommendations. The CCPO introduces automated procedures for EoL decision-making and overcomes challenges related to data silos and lack of semantic interoperability by utilising extensive product provenance information to provide accurate recommendations. This evaluation underscores CCPO's notable improvement over existing techniques, particularly in its ability to automate and simplify EoL decision-making procedures in circular supply chains. These results highlight CCPO's alignment with applied computing objectives like enhanced practical decision-making and seamless integration with existing systems.

\begin{table*}[h]
  \scriptsize
  \caption{SWRL Rules for EoL Decision-making}
  \label{tab:swrl}
  \begin{tabular}{lp{10cm}p{4cm}}
    \toprule
    \textbf{Name} & \textbf{Rule} & \textbf{Description}\\
    \midrule
    Rule1 & \raggedright \texttt{Product(?p)\^{}referenceServiceLife(?p,?r)\^{}actualServiceLife(?p,?a)\^{} swrlb:subtract(?diff,?r,?a)\^{} swrlb:lessThanOrEqual(?diff,1) -> atEoL(?p, true) } & Products reaching their EoL \\ 
    
    \rowcolor{lightgray}
    Rule2.i & \raggedright \texttt{Product(?p)\^{}\textbf{atEoL(?p,true)}\^{}hasHealthState(?p,green)\^{}actualServiceLife(?p,?a)\^{} referenceServiceLife(?p,?r)\^{}swrlb:subtract(?diff,?r,?a)\^{}swrlb:greaterThan(?diff,0) -> suggestedEoLRoute(?p,StrongReuseSuggestion)} & Strong reuse suggestion\\ 

    Rule2.ii &\raggedright \texttt{Product(?p)\^{}\textbf{atEoL(?p,true)}\^{}hasHealthState(?p, amber) ->  suggestedEoLRoute(?p, WeakReuse\_ConsiderRefurbishmentSoon)} & Weak reuse suggestion\\
    
    \rowcolor{lightgray}
    Rule2.iii& \raggedright \texttt{Product(?p)\^{}\textbf{atEoL(?p,true)}\^{}hasHealthState(?p,red) -> suggestedEoLRoute(?p,CannotReuseDueToPoorProductHealth)}& Cannot reuse\\

    Rule3.i & \raggedright \texttt{Product(?p)\^{}hasEoLStrategy(?p,?s) -> eolStrategyExists(?p,true)} & Check if Manufacturer's EoL strategy exists\\ 

    \rowcolor{lightgray}
    Rule3.ii & \raggedright \texttt{Product(?p)\^{}\textbf{atEoL(?p,true)}\^{}\textbf{suggestedEoLRoute(p,CannotReuseDueToPoorProductHealth)}\^{} \textbf{eolStrategyExists(?p,true)} -> suggestedEoLRoute(?p,FollowManufacturerEoLStrategy)} & Follow EoL handling strategy\\ 

    Rule3.iii & \raggedright \texttt{Product(?p)\^{}\textbf{atEoL(?p, true)}\^{}hasMarketDemand(?p, high)\^{}\textbf{suggestedEoLRoute(?p, CannotReuseDueToPoorProductHealth)} -> suggestedEoLRoute(?p,RecycleDueToHighMarketDemand)} & Recycle if there is a demand for recycled product.\\

    \rowcolor{lightgray}
    Rule3.iv & \raggedright \texttt{Product(?p)\^{}\textbf{atEoL(?p, true)}\^{}\textbf{suggestedEoLRoute(?p, CannotReuseDueToPoorProductHealth)}\^{}hasMarketDemand(?p, low) -> suggestedEoLRoute(?p, DoNotRecycleDueToLowDemand)} & Do not recycle if market demand is low\\

    Rule4 & \raggedright \texttt{Product(?p)\^{}\textbf{atEoL(?p, true)\^{}suggestedEoLRoute(?p, DoNotRecycleDueToLowDemand)} -> suggestedEoLRoute(?p, SendToLandfill)} & Send to landfill if conditions for reuse and recycle fail\\

    \bottomrule
  \end{tabular}
\end{table*}

\section{Conclusion} \label{conclusion}

This paper introduced and assessed the Circular Construction Product Ontology, a novel framework created to improve informed EoL decision-making in construction supply chains. CCPO uniquely integrates comprehensive product provenance data to enhance transparency, traceability, data connectivity and automated decision-making while aligning with circularity standards.

The architecture of the CCPO model combines a modular ontology structure with standardised vocabulary and data integration capabilities. The model successfully supports data exchange in decentralised systems and is adaptable to different construction products. This underscores CCPO’s practical impact on enhancing sustainability and reducing environmental impact in real-world construction projects, demonstrating its value as a robust tool for applied computing in the construction industry.

This work encompassed a thorough methodology which included domain analysis, a practical scenario development, competency questions formulation, model implementation, converting organisational SLAs and standards requirements into SWRL rules and a rigorous evaluation. Consultations with domain experts ensured the model's practicality and usability. The model was evaluated against task-based competency questions demonstrating its effectiveness in managing the complexities of circular supply chains. By leveraging advanced SPARQL queries and SWRL rules, CCPO enables automated and accurate EoL recommendations significantly improving upon traditional manual decision-making methods. This highlights CCPO's applicability in construction supply chains.

Nevertheless, the CCPO does have its constraints. Firstly, accurate decision-making relies on data availability from stakeholders. Essentially, CCPO's ability to support EoL management is dependent on establishing systems to ensure product data is persistent. Also, legislative bodies have a part to play in mandating supply chain actors to produce product data according to given standards. Secondly, the need for advanced reasoning capabilities, though essential, increases the model's computational complexity. This challenge is however acceptable because the CCPO has no requirement for real-time handling in EoL decision-making. In addition, while the model shows adaptability potential, its current implementation is specifically tailored to construction products. To extend its relevance across other industries, further domain-specific customizations will be necessary. This targeted approach ensures that CCPO remains highly effective within its intended scope.

Future efforts include enhancing CCPO's efficiency by integrating machine learning algorithms to automate and refine decision-making processes for real-time applications. Endeavours in improving the CCPO's scalability when applied to large-scale data will also be explored. In addition, future work will explore the use of Multichain blockchain technology to secure provenance data and enhance transparency within the supply chain. Another interesting direction will be exploring how the CCPO's decision algorithm may be enhanced by more complex requirements for EoL handling in different domains. Finally, continuous cooperation with industry stakeholders will be required to maximize the model's long-term impact and ensure it stays aligned with evolving standards and technological advancements.

\section*{Acknowledgments}
 This work is supported in part by the Engineering and Physical Sciences Research Council "Digital Economy" programme: EP/V042521/1 and EP/V042017/1. We extend our thanks to the principal investigators and all team members. Notably, we are grateful to Dr. Qian Li whose support ensured this project aligns with accepted industry standards. We would like to also express our gratitude to the EPSRC project's industry partners (FIBREE and Kognitive) who provided valuable insight for this work.  Special thanks to Dr. Xiang Xie for providing technical assistance and insight into the construction domain and ontology development best practices.  

\bibliographystyle{unsrtnat}
\bibliography{references}  






\end{document}